\documentclass[10pt,twocolumn,letterpaper]{article}

\usepackage{cvpr}              

\usepackage{graphicx}
\usepackage{amsmath}
\usepackage{amssymb}
\usepackage{booktabs}

\usepackage{booktabs}                                   
\usepackage{multirow}                                   
\usepackage{tablefootnote}                              
\usepackage[symbol]{footmisc}                           

\usepackage{color}                                      
\usepackage{pifont}

\newcommand{\kETAL}     {\emph{et al.}}
\newcommand{\bI}        {\textbf{I}}
\newcommand{\bQ}        {\textbf{Q}}
\newcommand{\bA}        {\textbf{A}}
\newcommand{\bH}        {\textbf{H}}
\newcommand{\bW}        {\textbf{W}}
\newcommand{\bF}        {\textbf{F}}
\newcommand{\bM}        {\textbf{M}}
\newcommand{\bY}        {\textbf{Y}}

\newcommand{\bv}        {\textbf{v}}
\newcommand{\bq}        {\textbf{q}}

\newcommand{\ba}        {\textbf{a}}
\newcommand{\bx}        {\textbf{x}}
\newcommand{\by}        {\textbf{y}}
\newcommand{\bb}        {\textbf{b}}
\newcommand{\bg}        {\textbf{g}}
\newcommand{\bh}        {\textbf{h}}

\newcommand{\bbf}        {\textbf{f}}
\newcommand{\bm}        {\textbf{m}}
\newcommand{\bs}        {\textbf{s}}
\newcommand{\br}        {\textbf{r}}
\newcommand{\bo}        {\textbf{o}}

\newcommand{\mR}        {\mathcal{R}}
\newcommand{\mL}        {\mathcal{L}}

\newcommand{\eat}[1]{}                                  

\usepackage{times}
\usepackage{epsfig}
\usepackage{graphicx}
\usepackage{amsmath}
\usepackage{amssymb}

\usepackage[pagebackref,breaklinks,colorlinks]{hyperref}

\usepackage[capitalize]{cleveref}
\crefname{section}{Sec.}{Secs.}
\Crefname{section}{Section}{Sections}
\Crefname{table}{Table}{Tables}
\crefname{table}{Tab.}{Tabs.}

\def\cvprPaperID{7045} 
\def\confName{CVPR}
\def\confYear{2022}

\begin{document}

\title{Multi-Clue Reasoning with Memory Augmentation for \\ Knowledge-based Visual Question Answering}

\author{Chengxiang Yin$^{1}$, Zhengping Che$^{2}$, Kun Wu$^{1}$, Zhiyuan Xu$^{2}$, Jian Tang$^{2}$\\
$^{1}$Syracuse University, $^{2}$Midea Group \\
}
\maketitle

\begin{abstract}
Visual Question Answering (VQA) has emerged as one of the most challenging tasks in artificial intelligence due to its multi-modal nature.
However, most existing VQA methods are incapable of handling Knowledge-based Visual Question Answering (KB-VQA), which requires external knowledge beyond visible contents to answer questions about a given image.
To address this issue, we propose a novel framework that endows the model with capabilities of answering more general questions, and achieves a better exploitation of external knowledge through generating \textbf{M}ultiple \textbf{C}lues for \textbf{R}easoning with \textbf{Mem}ory \textbf{N}eural \textbf{N}etworks (\textbf{MCR-MemNN}).
Specifically, a well-defined detector is adopted to predict image-question related relation phrases,
each of which delivers two complementary clues to retrieve the supporting facts from an external knowledge base (KB),
and these facts are further encoded into a continuous embedding space using a content-addressable memory.
%
Afterwards, mutual interactions between visual-semantic representation and the supporting facts stored in memory are captured to distill the most relevant information in three modalities (i.e., image, question, and KB).
Finally, the optimal answer is predicted by choosing the supporting fact with the highest score.
We conduct extensive experiments on two widely-used benchmarks.
The experimental results well justify the effectiveness of MCR-MemNN as well as its superiority over other KB-VQA methods.
%
\end{abstract}

\section{Introduction}
Recently, significant progress has been made for Visual Question Answering (VQA)~\cite{biten2019scene, chen2020counterfactual, gao2020multi, jiang2020defense, wang2020general},
which requires a joint comprehension of multi-modal content from images and natural language.
A VQA agent is expected to deliver the correct answer to a text-based question according to a given image.
Most of the existing VQA models~\cite{chen2015abc, lu2016hierarchical, ma2016learning, malinowski2015ask,  xiong2016dynamic} and datasets~\cite{krishna2017visual, malinowski2014multi, ren2015image, zhu2016visual7w} have focused on simple questions, which are answerable by solely analyzing the question and image, \textit{i.e., no external knowledge is required}.
However, a truly `AI-complete' VQA agent is required to combine both visual and semantic observations with external knowledge for reasoning, which is effortless for human but challenging for machine.
Therefore, to bridge the gap between human-level intelligence and algorithm design, Knowledge-based VQA (KB-VQA)~\cite{wang2017fvqa} is introduced to automatically find answers from the knowledge base (KB), 
given image-question pairs.

Some efforts have been made in this direction. 
%
To begin with, Wang~\kETAL~\cite{wang2017fvqa} presented a Fact-based VQA (FVQA) dataset to support a much deeper reasoning. FVQA consists of questions that require external knowledge to answer.
%
Several classical solutions~\cite{wang2017fvqa, wang2017explicit} have been proposed to solve FVQA by mapping each question to a query and retrieving supporting facts in KB through a keyword-matching manner. These supporting facts are processed to form the final answer.
%
However, these query-mapping based approaches with solely question parsing suffer from serious performance degradation when the information hint is not captured in the external KB or the visual concept is not exactly mentioned in the question.
Moreover, special information (i.e., visual concept type and answer source) should be determined in advance during querying and answering phases, which makes it hard to generalize to other datasets.
To address these issues, we introduce Multiple Clues for Reasoning, a new KB retrieval method, where a relation phrase detector is proposed to predict multiple complementary clues for supporting facts retrieval.

More recently, Out of the Box (OB)~\cite{narasimhan2018out} and Straight to Facts (STTF)~\cite{narasimhan2018straight} adopt cosine similarity technique to get the highest scoring fact for answer prediction, where the whole visual information (i.e., object, scene, and action) and the whole semantic information (i.e., all question words) are indiscriminately applied to infer the final answer by implicit reasoning. 
%
However, for an image-question pair, only part of the visual content in the image and several specific words in the question are relevant to a given supporting fact.
Additionally, the direct concatenation of image-question-entity embedding makes it hard to adaptively capture information among different modalities.
To exploit the inter-relationships among three modalities (i.e., image, question and KB), we propose a two-way attention mechanism with memory augmentation to model the interactions between visual-semantic representation and the supporting facts. Then the most relevant information in the three modalities can be distilled.
%
%
\begin{figure}[t!]
\centering
\includegraphics[width=1.0\columnwidth]{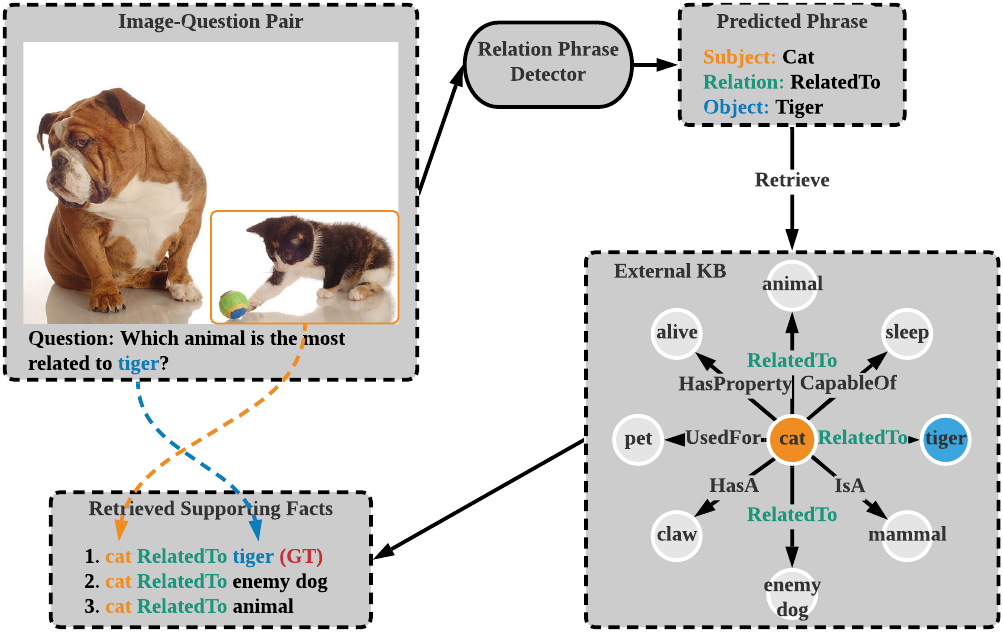}
\caption{An illustration of Multi-Clue Reasoning.}
\label{Fig::Motivation-MCR}
\end{figure}

In this paper, to deal with the KB-VQA task and address the aforementioned issues, we present a novel framework that focuses on achieving a better exploitation of external knowledge through \textit{generating \textbf{M}ultiple \textbf{C}lues for \textbf{R}easoning with \textbf{Mem}ory \textbf{N}eural \textbf{N}etworks}. 
We name it as \textbf{MCR-MemNN} for short.
Specifically, as illustrated by Figure~\ref{Fig::Motivation-MCR}, 
the image-question pair is encoded into a visual-semantic representation, 
which serves as the input to a well-defined detector to obtain a three-term relation phrase (i.e., \textit{$\langle$ Subject, Relation, Object $\rangle$}).
In this case, either the subject (i.e., cat) or the object (i.e., tiger) acts a clue to retrieve supporting facts in the external KB,
and both deliver a fact set including the ground-truth fact (i.e., cat RelatedTo tiger).
In this manner, the ground-truth fact can be successfully fetched as long as one of the two complementary clues is predicted.
Note that the predicted relation (i.e., RelatedTo) is adopted to filter out some facts with different relations.
The retained supporting facts, including the ground-truth, are further encoded into a continuous embedding space and stored in a content-addressable memory, where each memory slot corresponds to a supporting fact.
Afterwards, for reasoning procedure, we assume that the visual and semantic content in the image-question pair can contribute to a better exploitation of the external knowledge (i.e., supporting facts).
Analogously, the external knowledge is helpful for a better understanding of the image-question pair.
%
Therefore, we employ a two-way attention mechanism,
%
which is not only intended to focus on the important aspects of the memory in light of the image-question pair, but also the important parts of image and question in light of the memory.

The main contributions of this paper are summarized as follows:
(1) We demonstrate a new KB retrieval method with two complementary clues (i.e., subject and object), which makes it a lot easier to fetch the ground-truth supporting fact.
(2) A two-way attention mechanism with memory augmentation is employed to model the inter-relationships among image, question and KB to distill the most relevant information in the three modalities.
(3) We perform extensive experiments on two widely-used benchmarks, which shows that the proposed MCR-MemNN is an effective framework customized for KB-VQA.








\section{Related Work}
\subsection{Visual Question Answering}
Visual Question Answering (VQA) has emerged as a typical and popular multimedia application, which has been studied by quite a few recent works~\cite{anderson2018bottom, ma2018visual, teney2018tips, yin2019memory, zadeh2019social, yin2021hierarchical}.
A typical CNN-RNN based approach, Neural-Image-QA~\cite{malinowski2015ask}, learns an end-to-end trainable model by feeding both the image features and questions into LSTM.
Recently, Huang~\kETAL~\cite{huang2019multi} proposes a multi-grained attention mechanism, that addresses the failed cases on small objects or uncommon concepts through learning word-object correspondence.
%
%
%
To go a step further, graph neural network is leveraged in~\cite{hu2019language} to support complex relational reasoning among objects conditioned on the textual input.
Even though lots of methods and datasets have been proposed for VQA task, all of them are targeting at the simple questions, and none of them can deal with the general cases that external knowledge is required for a deep reasoning.

\subsection{Knowledge-based Visual Question Answering}
%
%
To deal with the knowledge-based visual question answering (KB-VQA) tasks, combination of visual observations with the external knowledge is required, which, however, is undoubtedly challenging.
In~\cite{wang2017fvqa}, Wang~\kETAL introduces a Fact-based VQA (FVQA) dataset, which requires and supports a much deeper reasoning with the help of the external knowledge bases.
It also provides a query-mapping based approach, which maps each question to a query to retrieve the supporting facts.
Similarly, Wang~\kETAL~\cite{wang2017explicit} transforms the question to an available query template and limits the question types.
However, those query-mapping based approaches suffers when the question does not focus on the most obvious visual concept, since the visual representation is not adopted for answer reasoning.
In some existing learning based approaches~\cite{narasimhan2018straight,narasimhan2018out}, both visual and semantic information are wholly provided and may introduce redundant information, which is harmful to the reasoning process.
As a solution, a heterogeneous graph neural network with multiple layers is employed in~\cite{zhu2020mucko} to collect the complementary evidence from three modalities under the guidance of question.
It achieves the state-of-the-art by adopting dense captioning for semantic parsing.
Analogously, this paper presents another solution by leveraging multiple clues for KB retrieval and leveraging a two-way attention mechanism with memory augmentation to model the inter-relationships among the three modalities including image, question and KB, which delivers a performance comparable with the state-of-the-art.

%

\section{Methodology}
We first describe the general definition and notations of Knowledge-based Visual Question Answering (KB-VQA), followed by the modeling of relation phrase detector.
Finally, we present the details about the proposed framework.
\subsection{Problem Statement}
\label{Sec::PS}
\begin{figure*}[t!]
\centering
\includegraphics[width=0.99\linewidth]{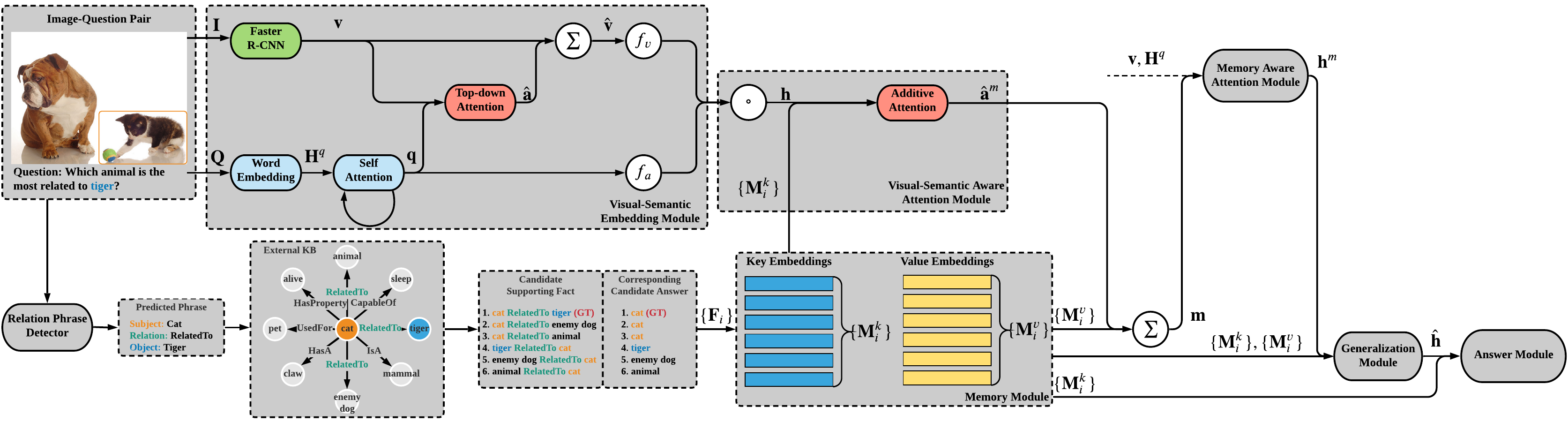}
\caption{Architecture of MCR-MemNN.}
\label{Fig::MCR-MemNN}
\end{figure*}

Given an image $\bI$ and a related question $\bQ$, the KB-VQA task aims to predict an answer $\bA$ from the external knowledge base (KB), which consists of facts in the form of triplet (i.e., \mbox{\textit{$\langle$ Subject, Relation, Object $\rangle$}}), where subject represents a visual concept, object represents an attribute or a visual concept and relation denotes the relationship between the subject and the object.
Note that the answer $\bA$ can be either the subject or object in the triplet.
The key of KB-VQA is to select the correct supporting fact and then determine the answer.
For convenience, in our notations, the fact \mbox{$\langle$ Subject, Relation, Object $\rangle$} corresponds to the answer subject; while its reversed form \mbox{$\langle$ Object, Relation, Subject $\rangle$} corresponds to the answer object.
For instance, as for the ground-truth supporting fact (i.e., cat \textit{RelatedTo} tiger) in Figure~\ref{Fig::MCR-MemNN}, the corresponding answer is cat, and the answer of its reversed form (i.e., tiger \textit{RelatedTo} cat) is tiger.
In this manner, during inference stage, the optimal answer is the first term of the predicted fact.

\begin{figure}[t!]
\centering
\includegraphics[width=1.00\columnwidth]{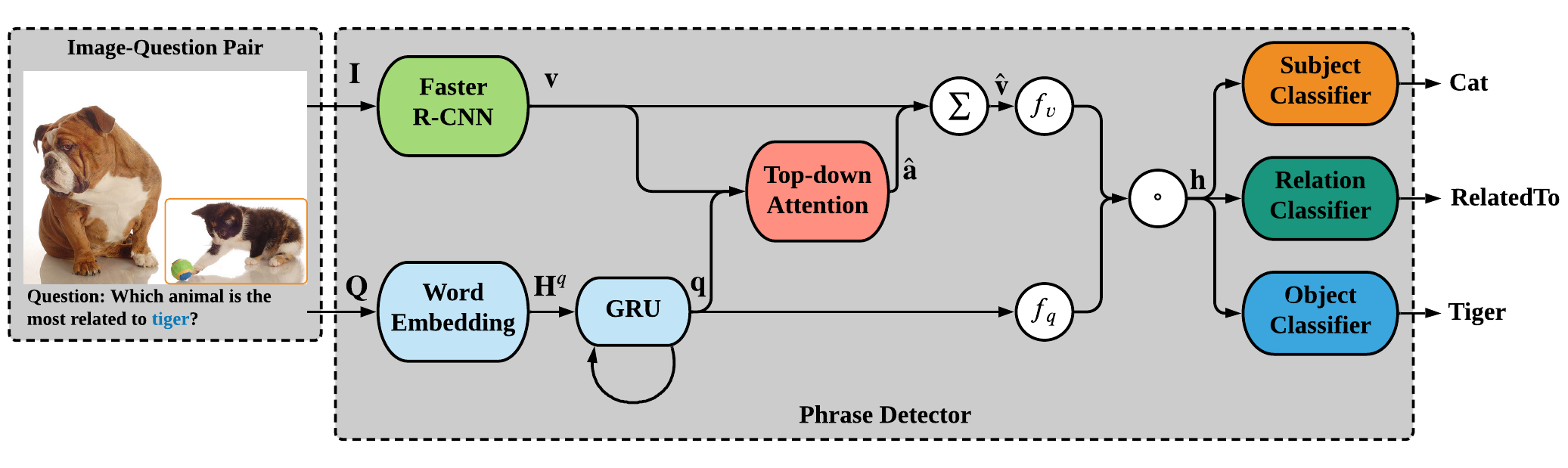}
\caption{Relation Phrase Detector.}
\label{Fig::RP-Detector}
\end{figure}

\subsection{Relation Phrase Detector}
Same as the fact in KB, the relation phrase is depicted in a form of triplet (i.e., \mbox{\textit{$\langle$ Subject, Relation, Object $\rangle$}}), which contains two complementary clues (i.e., subject and object) for KB retrieval.
As mentioned, the two complementary clues make it a lot easier to fetch the ground-truth supporting fact, and the relation predicted can be leveraged to filter out the facts with wrong relations.
Therefore, we develop a detector to obtain a relation phrase based on the visual and semantic content of the image-question pair.

As shown in Figure~\ref{Fig::RP-Detector}, given an image-question pair $\{\bI, \bQ\}$, the relation phrase prediction is formulated as a multi-task classification problem following~\cite{lu2018r}.
For image embedding, we use a Faster R-CNN~\cite{ren2015faster} in conjunction with the ResNet-101~\cite{he2016deep} pretrained on Visual Genome dataset~\cite{krishna2017visual} to generate an output set of image features $\bv$ as the visual representation of the input image.
\begin{equation}
\bv = \mbox{Faster R-CNN}(\bI)
\label{Eq:eq1}
\end{equation}

\noindent where $\bv \in {\mR}^{2048 \times K}$ is based on the bottom-up attention~\cite{anderson2018bottom}, which represents the ResNet features centered on the top-$K$ objects in the image. $K$ is set to 36 in our experiments.

For question embedding, we first transfer each word in $\bQ$ into a feature vector using the pre-trained 300D GloVe~\cite{pennington2014glove} vectors, and use randomly initialized vectors for the words which are out of GloVe's vocabulary.
Here we denote the resulting sequence of word embeddings as ${\bH}^{q}$.
Then the Gated Recurrent Unit (GRU)~\cite{chung2014empirical} with hidden state dimension 512 is adopted to encode the ${\bH}^{q}$ as semantic representation $\bq \in {\mR}^{512}$.
\begin{equation}
\bq = \mbox{GRU}({\bH}^{q})
\label{Eq:eq2}
\end{equation}

To encode the image and question in a shared embedding space, top-down attention~\cite{anderson2018bottom} is employed to fuse the visual representation $\bv$ and semantic representation $\bq$.
Specifically, for each object from $i = 1 ... K$, its feature ${\bv}_{i}$ is concatenated with the semantic representation $\bq$, and then passed through a non-linear layer $f_a$ and a linear layer to obtain its corresponding attention weight $a_{i}$.
\begin{align}
a_{i} & = {\bW}_{a}{f_a}([{\bv}_{i}, \bq]) \\
\hat{\ba} & = \mbox{softmax}(\ba) \\
\hat{\bv} & = {\sum}_{i = 1}^{K} \hat{a}_{i} {\bv}_{i}
\label{Eq:eq3}
\end{align}

\noindent where ${\bW}_{a}$ is a learnable weight vector and the attention weights $\ba$ are normalized with a softmax function to $\hat{\ba}$. The image features are weighted by the normalized attention values to get the weighed visual representation $\hat{\bv} \in {\mR}^{2048}$. Following~\cite{teney2018tips}, the non-linear layer $f_a: \bx \in {\mR}^m \to \by \in {\mR}^n$ with parameters $a$ is defined as follows:
\begin{align}
\widetilde{\by} & = \mbox{tanh} (\bW \bx + \bb) \\
\bg & = \sigma ({\bW}^{'} \bx + {\bb}^{'}) \\
\by & = \widetilde{\by} \circ \bg
\label{Eq:eq4}
\end{align}

\noindent where $\sigma$ is the sigmoid activation function, $\bW, {\bW}^{'} \in {\mR}^{n \times m}$ and $\bb, {\bb}^{'} \in {\mR}^{n}$ are learnable parameters, and $\circ$ denotes the element-wise multiplication.

A joint embedding $\bh$ of the question and the image is obtained by the fusion of the weighed visual representation $\hat{\bv}$ and the semantic representation $\bq$.
\begin{equation}
\bh = f_{v}(\hat{\bv}) \circ f_{q}(\bq)
\label{Eq:eq5}
\end{equation}

\noindent where $\bh \in {\mR}^{512}$. Both $f_{v}$ and $f_{q}$ are the non-linear layers with the same form as $f_{a}$. The joint embedding $\bh$ is then fed into a group of linear classifiers for the prediction of subject, relation and object in a relation phrase.
%
\begin{align}
\hat{\bs} & = \mbox{softmax}({\bW}_{s} \bh + {\bb}_{s}) \\
\hat{\br} & = \mbox{softmax}({\bW}_{r} \bh + {\bb}_{r}) \\
\hat{\bo} & = \mbox{softmax}({\bW}_{o} \bh + {\bb}_{o})
\label{Eq:eq6}
\end{align}

\noindent where ${\bW}_{s}, {\bW}_{r}, {\bW}_{o}$ and ${\bb}_{s}, {\bb}_{r}, {\bb}_{o}$ are learnable parameters, and
$\hat{\bs}, \hat{\br}, \hat{\bo}$
denote the predicted probability for subject, relation and object over each candidate, respectively.
The loss function for the relation phrase detector is defined as
\begin{equation}
{\mL}_{d} = {\lambda}_{s} {\mL}_{c}(\bs, \hat{\bs}) + {\lambda}_{r} {\mL}_{c}(\br, \hat{\br}) + {\lambda}_{o} {\mL}_{c}(\bo, \hat{\bo})
\label{Eq:eq7}
\end{equation}

\noindent where
$\bs, \br, \bo$ are the ground-truth labels for subject, object and relation, respectively.
${\mL}_{c}$ represents the cross-entropy loss, and the weights of the loss terms (i.e., ${\lambda}_{s}, {\lambda}_{r}, {\lambda}_{o}$) are set to 1.0 in our experiments.

\subsection{MCR-MemNN}
As shown in Figure~\ref{Fig::MCR-MemNN}, besides the Relation Phrase Detector, our proposed MCR-MemNN framework consists of six components, which are {Memory Module}, {Visual-Semantic Embedding Module}, {Visual-Semantic Aware Attention Module}, {Memory Aware Attention Module}, {Generalization Module} and {Answer Module}.

\subsubsection{Memory Module}

For an image-question pair $\{\bI, \bQ\}$, after the three-term phrase (i.e., Subject, Relation, Object) is predicted by the relation phrase detector, all the facts in the external KB with entities pointed by the subject or pointed to the object within $h$ hops are collected as candidate supporting facts. $h$ is set to 1 in our experiments. The relation predicted is leveraged to further filter out the facts with wrong relations.
Then a set of candidate supporting facts $\{{\bF}_{i}\}_{i = 1}^{N}$ is collected, where $N$ denotes the number of facts in the fact set, and each fact ${\bF}_{i}$ consists of a sequence of words.

Afterwards, similar to question embedding, each fact ${\bF}_{i}$ is transformed to a sequence of word embeddings ${\bH}^{f}_{i}$ based on GloVe's vocabulary, and encoded using a BiLSTM to get its representation ${\bbf}_{i} \in {\mR}^{128}$.
\begin{equation}
{\bbf}_{i} = \mbox{BiLSTM}({\bH}^{f}_{i})
\label{Eq:eq8}
\end{equation}

To store the candidate supporting facts, a key-value structured memory network~\cite{miller2016key} is leveraged. The key memory is used in the addressing stage, while the value memory is used in the reading stage.
The representation ${\bbf}_{i}$ of each fact is passed through two separate linear layers to obtain its key embedding ${\bM}^{k}_{i} \in {\mR}^{128}$ and value embedding ${\bM}^{v}_{i} \in {\mR}^{128}$ respectively.
\begin{align}
{\bM}^{k}_{i} & = {\bW}_{k} {\bbf}_{i} \\
{\bM}^{v}_{i} & = {\bW}_{v} {\bbf}_{i}
\label{Eq:eq9}
\end{align}

\noindent where ${\bW}_{k} \in {\mR}^{128 \times 128}$ and ${\bW}_{v} \in {\mR}^{128 \times 128}$ are learnable parameters.
For the set of candidate supporting facts $\{{\bF}_{i}\}_{i = 1}^{N}$, we have a set of key embeddings ${\bM}^{k} = \{{\bM}^{k}_{i}\}_{i = 1}^{N}$ and a set of value embeddings ${\bM}^{v} = \{{\bM}^{v}_{i}\}_{i = 1}^{N}$.
Note that one memory slot is defined as a pair of key embedding and value embedding (i.e., $\{ {\bM}^{k}_{i}, {\bM}^{v}_{i} \}$) of one candidate supporting fact.

\subsubsection{Visual-Semantic Embedding Module}
The visual-semantic embedding module has basically the same architecture as the relation phrase detector and the only difference is about the question embedding before top-down attention, where self-attention are applied over the sequence of word embeddings ${\bH}^{q}$ to get the semantic representation $\bq \in {\mR}^{128}$.
\begin{align}
{\hat{\ba}^{qq}} & = \mbox{softmax}({({\bH}^{q})}^{\intercal} {\bH}^{q}) \\
{\bq} & = \mbox{BiLSTM} ([{\bH}^{q} {({\hat{\ba}}^{qq})}^{\intercal}, {\bH}^{q}])
\label{Eq:eq10}
\end{align}

\noindent where the ${\bH}^{q}$ are weighted by the normalized values, concatenated with itself and fed into a BiLSTM.
Afterwards, same procedures are conducted to obtain the visual-semantic representation $\bh \in {\mR}^{128}$.

\subsubsection{Visual-Semantic Aware Attention Module}
\label{Sec::VS-AA}

Given visual-semantic representation $\bh$, we apply an attention over all the memory slots $\{ {\bM}^{k}_{i}, {\bM}^{v}_{i} \}_{i = 1}^{N}$ to obtain the memory summary $\bm \in {\mR}^{128}$ in light of the visual and semantic content of the image-question pair.
\begin{align}
a_{i}^{m} & = {\bW}_{3} \mbox{tanh} ({\bW}_{1} \bh + {\bW}_{2} {\bM}_{i}^{k}) \\
{\hat{\ba}}^{m} & = \mbox{softmax} ({\ba}^{m}) \\
\bm & = {\sum}_{i = 1}^{N} \hat{a}_{i}^{m} {\bM}_{i}^{v}
\label{Eq:eq11}
\end{align}

\noindent where ${\bW}_{1} \in {\mR}^{128 \times 128}, {\bW}_{2} \in {\mR}^{128 \times 128}, {\bW}_{3} \in {\mR}^{1 \times 128}$ are learnable parameters.
The attention for each memory slot is calculated and normalized based on $\bh$ and the corresponding key embedding ${\bM}^{k}_{i}$.
Then the set of value embeddings ${\bM}^{v} = \{{\bM}^{v}_{i}\}_{i = 1}^{N}$ are weighted to get the memory summary $\bm$.

\subsubsection{Memory Aware Attention Module}
\label{Sec::M-AA}

As we have obtained the memory summary $\bm$, we proceed to compute the attentions over all the question words and all the image features in light of the memory.

Given memory summary $\bm$, the sequence of word embeddings ${\bH}^{q}$ and the set of image features $\bv$,
the memory-aware question embedding ${\bq}^{m} \in {\mR}^{128}$ and the memory-aware image embedding ${\bv}^{m} \in {\mR}^{2048}$
are derived as follows:
\begin{align}
{\hat{\ba}}^{q} & = \mbox{softmax}({({\bH}^{q})}^{\intercal} \bm) \\
{\bq}^{m} & = {\bH}^{q} {\hat{\ba}}^{q} \\
{\hat{\ba}}^{v} & = \mbox{softmax}({({\bW}_{v} \bv)}^{\intercal} \bm) \\
{\bv}^{m} & = \bv {\hat{\ba}}^{v}
\label{Eq:eq12}
\end{align}

\noindent where ${\hat{\ba}}^{q}$ represents the normalized memory aware attention over all the question words, ${\hat{\ba}}^{v}$ represents the normalized memory aware attention over all the image features and ${\bW}_{v} \in {\mR}^{128 \times 2048}$ are learnable parameters.

The visual-semantic representation ${\bh}^{m}$ in light of the memory is obtained by the fusion of the memory-aware question embedding ${\bq}^{m}$ and the memory-aware image embedding ${\bv}^{m}$.
\begin{equation}
{\bh}^{m} = f_{v}^{m}({\bv}^{m}) \circ f_{q}^{m}({\bq}^{m})
\label{Eq:eq13}
\end{equation}

\noindent where ${\bh}^{m} \in {\mR}^{128}$. Both $f_{v}^{m}$ and $f_{q}^{m}$ are the non-linear layers with the same form as $f_{a}$.
Note that the aforementioned two-way mechanism corresponds to Sections~\ref{Sec::VS-AA} and~\ref{Sec::M-AA}.

\subsubsection{Generalization Module}

Inspired by~\cite{chen2019bidirectional}, another hop of the attention process is conducted over the memory before answer prediction.
Attention mechanism of Section~\ref{Sec::VS-AA} is leveraged here, given memory aware visual-semantic representation ${\bh}^{m}$, to fetch the most relevant information ${\bm}^{h} \in {\mR}^{128}$ from the memory to obtain the final visual-semantic representation $\hat{\bh} \in {\mR}^{128}$.
To be more specific, the fetched information ${\bm}^{h}$ is concatenated with ${\bh}^{m}$ and fed into a GRU to update the visual-semantic representation.
To the end, a batch normalization (BN) layer is used.
\begin{align}
{a}_{i}^{h} & = {\bW}_{6} \mbox{tanh} ({\bW}_{4} {\bh}^{m} + {\bW}_{5} {\bM}_{i}^{k}) \\
\hat{\ba}^{h} & = \mbox{softmax} ({\ba}^{h}) \\
{\bm}^{h} & = {\sum}_{i = 1}^{N} \hat{a}_{i}^{h} {\bM}_{i}^{v} \\
\hat{\bh} & = \mbox{BN} ({\bh}^{m} + \mbox{GRU} ([{\bh}^{m}, {\bm}^{h}]))
\label{Eq:eq14}
\end{align}

\noindent where ${\bW}_{4} \in {\mR}^{128 \times 128}, {\bW}_{5} \in {\mR}^{128 \times 128}, {\bW}_{6} \in {\mR}^{1 \times 128}$ are learnable parameters.

\subsubsection{Answer Module}

Given the final visual-semantic representation $\hat{\bh}$, and the set of key embeddings ${\bM}^{k} \in {\mR}^{128 \times N}$,
the key embedding ${\bM}_{i}^{k}$ of each candidate supporting fact is concatenated with the $\hat{\bh}$ to compute the probability of whether the fact is correct.
The predicted supporting fact is
\begin{equation}
\mbox{argmax}_{i} \mbox{softmax}({\bW}_{7}[\hat{\bh}, {\bM}^{k}] + {\bb}^{7})
\label{Eq:eq15}
\end{equation}

\noindent where $i = 1, ..., N$, ${\bW}_{7} \in {\mR}^{1 \times 256}$ and ${\bb}^{7} \in {\mR}^{N}$ are learnable parameters.
As we have stated in Section~\ref{Sec::PS}, the optimal answer is the first term of the predicted fact.

\subsubsection{Loss Function}

Once we have the candidate supporting facts retrieved from the external KB, all the learnable parameters of the proposed MCR-MemNN (besides the Relation Phrase Detector) are trained in an end-to-end manner by minimizing the following loss function over the training set.
\begin{equation}
\mL = \frac{1}{D} \sum_{k = 1}^{D} {\mL}_{c}({\bY}_{k}, \hat{\bY}_{k})
\label{Eq:eq16}
\end{equation}
%

\noindent where ${\mL}_{c}$ is defined as the cross-entopy loss, $D$ is the number of training samples, ${\bY}_{k}$ and $\hat{\bY}_{k}$ represent the ground-truth label and the predicted probability over each candidate supporting fact.


%
\section{Performance Evaluation}
We employed two compelling benchmarks, FVQA~\cite{wang2017fvqa} and Visual7W+ConceptNet~\cite{li2017incorporating} to evaluate the proposed MCR-MemNN on Knowledge-based Visual Question Answering (KB-VQA) task.

\subsection{Benchmark Datasets}
\noindent \textbf{FVQA.} The Factual Visual Question Answering (FVQA) dataset consists of $2,190$ images, $5,286$ questions and $4,126$ unique facts corresponding to the questions.
The external KB of FVQA, consisting of $193,449$ facts, are constructed based on three structured KBs, including DBpedia~\cite{auer2007dbpedia}, ConceptNet~\cite{liu2004conceptnet} and WebChild~\cite{tandon2014acquiring}.
Following~\cite{wang2017fvqa}, the top-1 and top-3 accuracies are averaged over five test splits.

\noindent \textbf{Visual7W+ConceptNet.} The Visual7W+ConceptNet dataset, built by~\cite{li2017incorporating}, is a collection of knowledge-based questions with images sampled from the test split of Visual7W~\cite{zhu2016visual7w} dataset.
It consists of $16,850$ open domain question-answer pairs based on $8,425$ images.
Note that the supporting facts of each question-answer pair are retrieved directly from the ConceptNet, which serves as the external KB.
Following~\cite{wang2017fvqa}, the top-1 and top-3 accuracies are calculated over the test set.

\subsection{Experimental Setup}
\subsubsection{Implementation Details}
For the training of the relation phrase detector, the model was trained with Adam optimizer~\cite{kingma2014adam} with an initial learning rate of $1 \times 10^{-4}$ and weight decay of $1 \times 10^{-6}$, and the batch size is set to 32.

For the training of the proposed MCR-MemNN (besides the relation phrase detector), the model was trained with Adam optimizer with an initial learning rate of $1 \times 10^{-3}$ and weight decay of $1 \times 10^{-6}$, and the batch size is set to 64.
Top 40 predicted subjects and objects were adopted as clues to retrieve candidate supporting facts from the external KB.
The memory size $N_{mem}$ was set to 96.
Note that no matter the number of candidate supporting facts $N$ is larger than, equal to or smaller than $N_{mem}$, the ground-truth fact is preserved and the negative facts are randomly selected to fill up the remaining memory.
Our code was implemented in PyTorch~\cite{paszke2017automatic} and run with NVIDIA Tesla P100 GPUs.

\subsubsection{Baselines}
For the FVQA dataset, CNN-RNN based approaches including LSTM-Question+Image+Pre-VQA~\cite{wang2017fvqa} and Hie-Question+Image+Pre-VQA~\cite{wang2017fvqa},
semantic parsing based approaches including FVQA (top-3-QQmaping)~\cite{wang2017fvqa} and FVQA (Ensemble)~\cite{wang2017fvqa},
learning-based approaches including Straight to the Facts (STTF)~\cite{narasimhan2018straight}, Out of the Box (OB)~\cite{narasimhan2018out}, Reading Comprehension~\cite{li2019visual}, and Multi-Layer Cross-Modal Knowledge Reasoning (Mucko)~\cite{zhu2020mucko} are compared with the proposed MCR-MemNN.
%
%

For the Visual7W+ConceptNet dataset, learning based approaches including KDMN~\cite{li2017incorporating}, Out of the Box (OB)
are compared with the proposed MCR-MemNN.
Note that both KDMN and MCR-MemNN adopted memory augmentation technique for storing the retrieved external knowledge.

\begin{table}[h]
\centering
\caption{Results for the relation phrase detector on FVQA.}
\label{Tab::RPD}
\resizebox{0.99\columnwidth}{!}{
\begin{tabular}{c|ccc|cc|cc}
\multirow{2}{*}{\textbf{Case}} &  \multicolumn{3}{c}{\textbf{Classification Acc.}} & \multicolumn{2}{c|}{\textbf{Recall}} & \multicolumn{2}{c}{\textbf{QA Acc.}} \\
 & \textbf{Sub} & \textbf{Obj} & \textbf{Union} & \textbf{w/ Rel} & \textbf{w/o Rel} & \textbf{w/ Rel} & \textbf{w/o Rel} \\
\midrule
\textbf{Top-20} & 64.31 & 39.43 & 69.80 & 78.56 & 82.08 & 60.36 & 62.76 \\
\textbf{Top-30} & 69.46 & 43.30 & 74.24 & 83.61 & 86.90 & 66.34 & 65.28 \\
\textbf{Top-40} & 72.65 & 45.71 & 76.22 & 85.76 & 88.85 & \textbf{70.92} & \textbf{68.85} \\
\textbf{Top-60} & 73.59 & 46.23 & 77.10 & 86.58 & 89.81 & 68.52 & 67.30 \\
\textbf{Top-100} & 75.60 & 47.61 & 79.37 & 88.21 & 90.33 & 68.02 & 65.57
\end{tabular}}
\end{table}

\subsection{Experimental Results}
\subsubsection{Relation Phrase Detector}
As shown in Table~\ref{Tab::RPD}, to evaluate the performance of the relation phrase detector, five different cases are considered, including Top-20, Top-30, Top-40, Top-60 and Top-100.
%
%
For each case, the classification accuracies of subject (i.e., `Sub'), object (i.e, `Obj') and both union (i.e., `Union') are calculated.
In addition, the answer recall is reported with the top-3 relation limitation (i.e., `w/~Relation') or not (i.e., `w/o~Relation').
Finally, the downstream question answering accuracy (`QA~Acc.') is also calculated.

For more clarity, the `Sub' in the case of Top-40 represents the fraction of test samples that the ground-truth subject is included in the
top 40 predicted subjects,
and these subjects are further adopted as the clues for KB retrieval.
The Recall w/~Relation in the case of Top-40 represents the fraction of test samples that the correct answer is included in the candidate answer set corresponding to the supporting facts retrieved by the top-40 subject or object clues and filtered by the top 3 predicted relations.
Note that the top-3 classification accuracy for relation prediction using the relation phrase detector is $93.20\%$.

Results in Table~\ref{Tab::RPD} show that both union accuracy and answer recall of the Top-40 case are higher than those of the Top-20 case, and improvements of $10.56\%$ and $6.09\%$ on downstream QA accuracies (with and without relation filtering) are caused.
Even though much more relevant supporting facts are retrieved from the external KB in the Top-100 case, which delivers higher both union accuracy and answer recall, the downstream QA accuracies are lower than those of the Top-40 case.
This observation clearly shows that excessive retrieved facts can lead to more redundant information, which is harmful to the reasoning process.
We choose the top 40 subjects and objects as clues for KB retrieval as this gives the best downstream QA accuracies.
\begin{table}[h]
\centering
\caption{Accuracy comparison on FVQA.}
\label{Tab::Acc-FVQA}
\resizebox{0.95\columnwidth}{!}{
\begin{tabular}{c|cc}
\multirow{2}{*}{\textbf{\begin{tabular}[c]{@{}c@{}}Method\end{tabular}}} & \multicolumn{2}{c}{\textbf{Overall Accuracy}} \\
 & \textbf{top-1} & \textbf{top-3} \\
\midrule
LSTM-Question+Image+Pre-VQA~\cite{wang2017fvqa} & 24.98 & 40.40 \\
Hie-Question+Image+Pre-VQA~\cite{wang2017fvqa} & 43.14 & 59.44 \\
FVQA (top-3-QQmaping)~\cite{wang2017fvqa} & 56.91 & 64.65 \\
FVQA (Ensemble)~\cite{wang2017fvqa} & 58.76 & - \\
Straight to the Facts (STTF)~\cite{narasimhan2018straight} & 62.20 & 75.60 \\
Reading Comprehension~\cite{li2019visual} & 62.94 & 70.08 \\
Out of the Box (OB)~\cite{narasimhan2018out} & 69.35 & 80.25 \\
Mucko (w/o Semantic Graph)~\cite{zhu2020mucko} & \textbf{71.28} & \textbf{82.76} \\
\midrule
\textbf{MCR-MemNN (Ours)} & \textbf{70.92} & \textbf{81.83}
\end{tabular}}
\end{table}
\begin{table}[h]
\centering
\caption{Accuracy comparison on Visual7W+ConceptNet.}
\label{Tab::Acc-Vis7W}
\resizebox{0.8\columnwidth}{!}{
\begin{tabular}{c|cc}
\multirow{2}{*}{\textbf{Method}} & \multicolumn{2}{c}{\textbf{Overall Accuracy}} \\
 & \textbf{top-1} & \textbf{top-3} \\
\midrule
KDMN-NoKnowledge~\cite{li2017incorporating} & 45.1 & - \\
KDMN-NoMemory~\cite{li2017incorporating} & 51.9 & - \\
KDMN~\cite{li2017incorporating} & 57.9 & - \\
KDMN-Ensemble~\cite{li2017incorporating} & 60.9 & - \\
Out of the Box (OB)~\cite{narasimhan2018out} & 57.32 & 71.61 \\
\midrule
\textbf{MCR-MemNN (Ours)} & \textbf{64.23} & \textbf{79.18}
\end{tabular}}
\end{table}
\begin{figure*}[t!]
\centering
\includegraphics[width=0.99\linewidth]{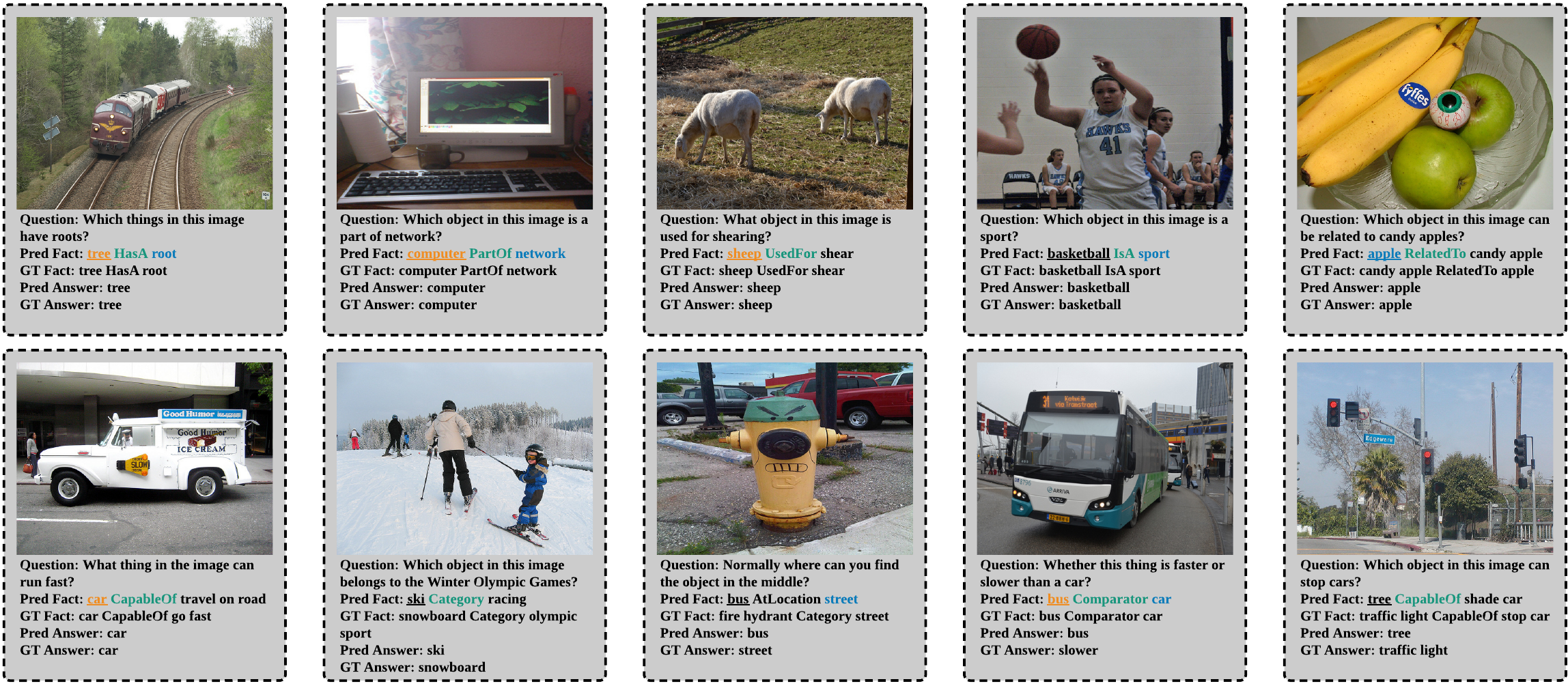}
\caption{Success and failure cases.
(Terms in yellow/blue indicates the correctly predicted subjects/objects (in top-40), terms in green denotes the correctly predicted relations (in top-3) and terms underlined represents the predicted answer.)
}
\label{Fig::Quali-R}
\end{figure*}

\subsubsection{MCR-MemNN}
Tables~\ref{Tab::Acc-FVQA} and~\ref{Tab::Acc-Vis7W} show the comparison of the proposed  MCR-MemNN with the above-mentioned baselines on FVQA and Visual7W+ConceptNet, respectively, and the following observations can be made:

1) On FVQA dataset, MCR-MemNN outperforms almost all the baselines in terms of top-1 and top-3 accuracies.
To be specific, MCR-MemNN outperforms the semantic parsing based approaches including FVQA (top-3-QQmaping) and FVQA (Ensemble), and achieves more than $12\%$ boost on top-1 accuracy and more than $15\%$ boost on top-3 accuracy.
%
In addition, compared with the typical learning based approaches including STTF and OB, which wholly introduce the visual and semantic information without selection,
MCR-MemNN gains an improvement by leveraging a two-way attention mechanism (i.e., Sections~\ref{Sec::VS-AA} and~\ref{Sec::M-AA}) to exploit the inter-relationships among image, question and KB and distill the most relevant information in each of the three modalities.

2) Even though a heterogeneous graph neural network with high complexity is employed in Mucko, the proposed MCR-MemNN can still deliver a comparable performance.
Note that the full model of Mucko leveraged dense captions~\cite{johnson2016densecap} as input for performance improvement. For fair comparison, the semantic graph for dense caption parsing is removed.

3) On Visual7W+ConceptNet dataset, MCR-MemNN consistently outperforms a series of models based on KDMN, which leverages a dynamic memory network to preserve the retrieved external knowledge.
Since both MCR-MemNN and KDMN adopted the memory augmentation technique, this observation further evidences the effectiveness of modeling inter-relationships among image, question and KB for KB-VQA task.
%
Note that Mucko does not provide the result without Semantic Graph.
\begin{table}[h]
\centering
\caption{Ablation studies on FVQA. (Sub: Subject as Clue; Obj: Obj as Clue; Rel: Relation Filtering; Att: Two-way Attention)}
\label{Tab::Abl-FVQA}
\resizebox{0.85\columnwidth}{!}
{
\begin{tabular}{c|c|c|c|c|cc}
\multirow{2}{*}{\textbf{Case}} & \multirow{2}{*}{\textbf{Sub}} & \multirow{2}{*}{\textbf{Obj}} & \multirow{2}{*}{\textbf{Rel}} & \multirow{2}{*}{\textbf{Att}} & \multicolumn{2}{c}{\textbf{Overall Accuracy}} \\
 &  &  &  &  & \textbf{top-1} & \textbf{top-3} \\
\midrule
1 & \ding{51} &  &  &  & 61.60 & 68.18 \\
2 &  & \ding{51} &  &  & 44.07 & 52.72 \\
3 & \ding{51} & \ding{51} &  &  & 65.04 & 73.58 \\
4 & \ding{51} & \ding{51} & \ding{51} &  & 67.52 & 76.48 \\
5 & \ding{51} & \ding{51} &  & \ding{51} & 68.85 & 78.37 \\
\midrule
6 & \ding{51} & \ding{51} & \ding{51} & \ding{51} & \textbf{70.92} & \textbf{81.83}
\end{tabular}}
\end{table}

\subsubsection{Ablation Studies}
To validate the superiority of the proposed MCR-MemNN on KB-VQA, several ablation experiments were conducted based on FVQA dataset, and we can make the following observations based on Table~\ref{Tab::Abl-FVQA}:

1) MCR-MemNN adopts both subject and object as clues for KB retrieval, which leads to better performance.
Specifically, compared with Case-1, where only subject is adopted as clues for KB retrieval, there exists a jump on top-1 (i.e., $3.44\%$) and top-3 (i.e., $5.40\%$) accuracies when both of the subject and object are leveraged as clues in Case-3.
A similar gain can be observed when Case-3 is compared with Case-2.
%

2) To validate the effectiveness of the relation filtering, the experiment is conducted in Case-4, which achieves $2.48\%$ improvement over Case-3 on top-1 accuracy.
This observation clearly implies that the predicted relation can successfully remove the redundant supporting facts retrieved from the external KB.
Similarly, compared with Case-5 without relation filtering, Case-6 brings up an improvement of $2.07\%$.

3) The two-way attention mechanism (i.e., Sections~\ref{Sec::VS-AA} and~\ref{Sec::M-AA})
can deliver an additional performance gain on KB-VQA task. For instance, compared with Case-4, where the inter-relationships are not exploited among image, question and KB, Case-6 brings up improvements of $3.40\%$ on top-1 accuracy.
This indicates that the redundant information of the three modalities (i.e., image, question and KB) is removed during the reasoning process,
and the most relevant information is collected by modeling inter-relationships among the three modalities.

\subsubsection{Qualitative Results}
Figure~\ref{Fig::Quali-R} shows several success and failure cases using MCR-MemNN.
Two steps are indispensable to predict the correct answer:
(1) Either the correct subject or object is included in the top-40 predicted subject set or object set.
(2) The correct relation is predicted as the top-3 relations.
The ground-truth fact will be retrieved as one of the candidate supporting facts, if both the two steps are successfully accomplished.
For instance, all five samples in the first row have their corresponding ground-truth facts successfully predicted and the correct answers are delivered.
Specifically, the first two samples have both of their subjects and objects correctly predicted. The 3rd sample have its subject correctly predicted while the last two have their objects correctly predicted.
This clearly verifies the advantage of multiple clues reasoning, retrieval using two complementary clues (i.e., subject and object) makes it a lot easier to fetch the ground-truth fact and deliver the correct answer.
Some other cases are presented in the second row.
Generally, if a wrong fact is predicted (e.g., the 2nd, 3rd and 5th samples), the correct answer cannot be given.
However, in some special cases, even if the ground-truth fact is not successfully predicted (e.g., the 1st sample), the correct answer can still be delivered.
If the correct relation is not included in the top-3 (e.g., the 3rd sample), the correct answer cannot be given even if the subject or object is correctly predicted.
The MCR-MemNN always fails if the question is intend to know the relationship between subject and object (e.g., 4th sample).

\section{Conclusions}
This paper, by introducing \textbf{M}ultiple \textbf{C}lues for \textbf{R}easoning with \textbf{Mem}ory \textbf{N}eural \textbf{N}etwork (\textbf{MCR-MemNN}), 
presents a novel framework for knowledge-based visual question answering (KB-VQA).
Comprehensive experiments have been conducted on two widely-used benchmarks
and the extensive experimental results have shown that
1) Retrieval using two complementary clues (i.e., subject and object) makes it a lot easier to fetch the ground-truth fact and deliver the correct answer;
2) Two-way attention mechanism with memory augmentation can successfully model the inter-relationships among the three modalities including image, question and KB,
and brings up a remarkable performance gain on KB-VQA;
3) MCR-MemNN outperforms most of the KB-VQA methods and achieves a comparable performance with the state-of-the-art.

\clearpage

{\small
\bibliographystyle{ieee_fullname}
\bibliography{KB-VQA-CVPR2022}
}

\end{document}